# Ceasing hate with *MoH*: Hate Speech Detection in Hindi-English Code-Switched Language

Arushi Sharma[a,**], Anubha Kabra[b,**], Minni Jain[c,*]

[a] *Optum Global Advantage*
[b] *Adobe Inc.*
[c] *Delhi Technological University*

**Abstract**

Social media has become a bedrock for people to voice their opinions worldwide. Due to the greater sense of freedom with the anonymity feature, it is possible to disregard social etiquette online and attack others without facing severe consequences, inevitably propagating hate speech. The current measures to sift the online content and offset the hatred spread do not go far enough. One factor contributing to this is the prevalence of regional languages in social media and the paucity of language flexible hate speech detectors. The proposed work focuses on analyzing hate speech in Hindi-English code-switched language. Our method explores transformation techniques to capture precise text representation. To contain the structure of data and yet use it with existing algorithms, we developed *'MoH'* or (Map Only Hindi), which means 'Love' in Hindi. *'MoH'* pipeline which consists of language identification, Roman to Devanagari Hindi translitera-

[*] Corresponding author.
[**] Both authors contributed equally.
*URL:* arushi.net@outlook.com (Arushi Sharma), anukabra@adobe.com (Anubha Kabra), minnijain@dtu.ac.in (Minni Jain)

tion using a knowledge base of Roman Hindi words, and finally employs the fine-tuned Multilingual Bert, and MuRIL language models. We conducted several quantitative experiment studies on three datasets, and evaluated performance using Precision, Recall and F1 metrics. The first experiment studies *'MoH'* mapped text's performance with classical machine learning models and shows an average increase of 13% in F1 scores. The second compares the proposed work's scores with those of the baseline models and shows a rise in performance by 6%. Finally, the third compares the proposed *'MoH'* technique with various data simulations using the existing transliteration library. Here, *'MoH'* outperforms the rest by 15%. Our results demonstrate a significant improvement in the state-of-the-art scores on all three datasets.
*Keywords:* cyber hate, social media, data simulations, Bert, MuRIL, transfer learning, text classification, and machine learning

---

## 1. Introduction

Rapid digitalization has increased the prevalence of social media in today's world. More than half of the human population has a social identity, making up more than 3.6 billion people using the Internet (Clement, 2020). The Social Media platforms act as a global discussion forum due to their massive user base. While it creates a global unity and helps people of different cultures connect, social media is also widely used to spread cyberhate and cybercrimes (Baron, 2019). Hate speech is defined as the language that is used to express hatred towards a minority group or is intended to humiliate or insult the members of a group or an individual (Davidson et al., 2017). We propose a work that focuses on hate speech detection in a regional scope, with attention to the code-switched Hindi and English language.

This task is essential due to the following reasons:

*Harm caused by hate speech on social media.* Many social media posts tend to target or exploit an individual or a group with characteristics such as nationality, ethnicity, religion and sexual orientation. Such comments can cause the victims to develop inferiority complexes or completely damage their mental health. Furthermore, these may lead to cyber or even real-world crimes (Burnap & Williams, 2015; Zhang et al., 2018). In democratic countries, social media is widely used for political campaigns. During state elections, social media platforms have been used to influence people against communities and inculcate hate to take forward the political party's propaganda (Cadwalladr & Graham-Harrison, 2018).

*Prevalence of regional dialects.* It is easier for the masses to communicate and present their opinions in their native languages. Hence, a myriad of posts online is in regional expression. In 2020, the number of social media users was around 376 million, and this count is ever-rising (Keelery, 2020). Moreover, Hindi is the third-highest spoken language in the world, with around 500 million speakers (Ethnologue, 2020). The urban Hindi-speaking nations mix Hindi and English words and prefer using the Roman Hindi script (Hindi text written using the Latin alphabets) over the traditional Devanagari Hindi script for ease of communication (Das & Gambäck, 2015). A fair amount of work has been done to classify hate speech, narrowly in the scope of the English language (Arango et al., 2019). However, to reach the root of the problem, it is necessary to tackle the language barriers. Curbing hate speech online using high accuracy autonomous techniques has become the need of the hour.

There are numerous challenges in applications related to Hindi-English code-switched data:

1. Writing Hindi words using English characters results in numerous inconsistencies (Guha, 2015) in text as the Roman alphabets are used in multitude of ways while writing Hindi. For example, the Roman Hindi words **angoor (grape)** (अंगूर in Devanagari) and **alap(vocal music)** (आलाप in Devanagari), both include the letters ‘a’ in Roman script, but these letters have different representation in Devanagari Hindi. The ‘a’ in **angoor** is written as ‘अ’ while the ‘a’ in **alap** is written as ‘आ’ in Devanagari Hindi.
2. There is no standard Roman Hindi vocabulary (Mathur et al., 2018), hence different people represent the same Hindi word with different Roman spellings. For example, नमस्ते (Hindu Greeting) can be written as ‘namaste’, ‘namastey’, ‘numuste’ in Roman script. We will see how these multiple spellings affect model performance in Section 6. As a consequence of the non-standard Roman Hindi words representation, this is a low-resource text classification task.
3. Code-switching can use a variable grammatical structure. The atypical nature of the code-switching is attributed to the fact that there is no specific grammar structure followed (Mathur et al., 2018) i.e., the sentence structure can be that of English or Hindi depending on the user preference. This is evident in Table 1.

Table 1: The first row has an English Interrogative Sentence, and the second and third rows have the same first-row sentence written in code-switched English and Roman Hindi languages. The CS English row implies code-switching with English grammatical structure, and the CS Hindi row implies code-switching with Hindi grammatical structure.

| | |
|---|---|
| English: | *What's your excuse?* |
| CS English: | *What's your bahana?* |
| CS Hindi: | *Tumhara kya excuse hai?* |

Previous works rely on translating datasets into English to make use of the power of pre-trained language models, (Mathur et al., 2018) and (Bhat et al., 2015). However, these approaches suffer from syntactical and grammatical errors. Moreover, Previous works have explored feature extraction techniques in this task but reported poor results in differentiating between hate inducing and profane texts (Badjatiya et al., 2017), (Mathur et al., 2018).

To overcome the problems mentioned above, we introduce *MoH*, a word-level transliteration pipeline that identifies language tags, performs partial transliteration of code-switched text while preserving the semantic sense of the sentences. We further fine-tuned Multilingual Bidirectional Encoder Representations from Transformers (M-Bert) and Multilingual Representations for Indian Languages (MuRIL) for classification as this type of Devanagari Hindi-English code-switching is suitable for Bert based models (Pires et al., 2019). *MoH* means 'love' in Hindi and is an acronym for 'Map only Hindi', as we aim to skip English words and transliterate Roman Hindi into the standard writing script, i.e., Devanagari format. The proposed method does not change the sentence structure or use feature extraction; instead, it aims to correctly convert the Roman Hindi words into their Devanagari Hindi format. This method overcomes the shortcomings mentioned above in the following ways:

1. Overcoming word-level transliteration errors: We collected various databases (Roy et al., 2013; Kunchukuttan et al., 2017) to combine a vast knowledge base with word-to-word mapping and used Levenshtein similarity to convert Roman Hindi words correctly into Devanagari Hindi. (Refer section 5.)
2. Capturing the semantic structure using M-Bert and MuRIL: M-Bert can create suitable contextual embeddings for code-switched Devanagari Hindi and English words in a sentence without additional feature requirements (Pires et al., 2019). Hence using M-Bert embeddings curbs the structure-based problems. MuRIL is another Bert based model that is pre-trained on Indian languages. It is newly developed (Khanuja et al., 2021).

We perform extensive experimentation on three datasets to show that the proposed work has a significantly higher performance than the state-of-the-art methods. We compare our work with three lines of methodologies. First is a comparison of classical machine learning models used on raw as well as MoH mapped text. Next, we compare MoH mapped text on Bert based models with baseline works of the three datasets. Lastly, we perform a comparative study by transliterating text using Indic-Trans transliterator and training the MuRIL model.

The major contributions of our work are as follows:

1. This is the first work in Hate Speech Detection that preserves the semantic sense and syntactic structure of code switched text. The proposed MoH pipeline ensures that the words in code switched text are transliterated to their original languages without compromising on the semantic sense and syntactic structure of the text while assuring the correct spelling of each word. Moreover, we support our analysis with thorough experimental studies comparing our transliteration technique with a popular character-level translation library i.e: Indic-Trans as well as state-of-the-art feature extraction techniques.

2. This is the first work to fine-tune MuRIL (Khanuja et al., 2021) for Hate Speech Detection and provide a detailed comparison of scores between MuRIL and M-Bert.

3. To improve model performance, we proposed MoH text conversion steps. Their performance is measured by comparing them with a previous work that applied M-Bert on raw (without any conversion of transliterations) code switched text (Modha et al., 2020). The experimental results show that the proposed MoH pipeline performs significantly higher (15% higher F1 score), highlighting the performance boost that comes due to MoH text conversion steps.

4. The proposed method has a significant boost in performance over multiple state-of-the-art techniques on Precision, Recall, and F1 evaluation metrics. We record an average boost of 4% in F1 scores from the best performing baselines on three datasets (TRAC-I, HOT and HS).

5. We provide a detailed discussion and error analysis which gives an insight into a variety of texts that can be present on social media and how the proposed system has responded to them.

The organization of the paper is as follows: we provide the background of key concepts in Section 2, related work in Section 3, data collection and pre-processing in Section 4, and model pipeline in the Section 5. Next, we conduct several sets of experiments, shown in Section 6 to evaluate our model's performance with classical machine learning models, existing baselines, and Indic-Trans data simulations. We also show a detailed error analysis based on the results achieved. Lastly, in Section 7 we formulate inferences from the errors in our model predictions.

## 2. Background

In this section, we provide a brief background of all the key concepts used in our work.

### *2.1. Hindi and its representation in Roman Script*

Hindi is formally written using the Devanagari script, which has 33 consonants and 13 vowels, instead of Hindi written using the Roman script with 21 consonants and 5 vowels. Hence, the same Hindi word can be represented with multiple spellings when written in Roman script. Therefore, the social media text does not follow a standard transliteration. Moreover, the simple sentences in Hindi are of the form Subject + Object + Verb, instead of the Subject + Verb + Object structure in English. This is why, even though the second and third rows in Table 1 are code-switched text written in Roman script, they are different from each other in terms of their structure.

### *2.2. Hate speech terminology and scope*

The scope of hate speech in this manuscript is defined as:

Table 2: The following are examples showcasing the exclusivity between hate speech and profane text. Here, HS refers to hate speech. The subject is in italics, and profane words are in bold.

| | Post | Translation | Profane | HS |
|---|---|---|---|---|
| 1 | saari *ladkiyan* **pagal** hoti hai. Bilkul useless. | All *girls* are mad. Completely useless. | Yes | Yes |
| 2 | *Bangaldeshi* apne desh vapas jaa. | *Bangladeshi*, go back to your country. | No | Yes |
| 3 | *Rahul* bhi alag hi **g**du** hai. Dusro ke liye kuch bhi karta hai . | *Rahul* is an **a*s**, he is always ready to help others. | Yes | No |

1. Text that attacks an individual or community instigates hatred, with or without using profanity, qualifies as hate speech. For instance, in Table 2, the first two posts are humiliating specific individuals (shown in *italics*) with and without using profanity, respectively, and hence these posts are hate speech. The third post has a profane word (g**du), but it complements the subject (Rahul). Hence it is not hate speech.
2. The proposed work is focused narrowly on textual content/posts on social platforms.

Literary works have discussed the terminology of hate speech as vulgar or offensive language (Xiang et al., 2012). Now researchers evaluate and include many forms of expressions which spread, incite or justify hatred, violence, and discrimination. Hate speech has been analyzed in the form of cyberbullying in multiple works (Zhong et al., 2016; Van Hee et al., 2015). Furthermore, 'Othering language', another form of hate, signals a dichotomy between 'us' and 'them' (e.g. 'send them home'), hence alienating diverse cultural beings (Burnap et al., 2014; Alorainy et al., 2019).

### *2.3. Type of code-switching*

Three forms of code-switching alternatively employ English and Hindi sentences. *Inter-sentential code-switching* (Wei, 2020) comprises of sentences or clauses that are completely written in a language and then followed by sentences written in another language. For example, *Oh my god! yeh kya ho gya!*, where *yeh kya ho gya* means *what has happened.* Here, the first sentence is in English, but it is written in Roman Hindi. *Intra-sentential code-switching* (Wei, 2020) mixes the two languages in-between sentences. For example, *bahar accha weather hai.* which means 'The weather outside is nice today.' Here, the user inserted the English word *'weather'* into the grammatical structure of the Roman Hindi sentence, which means that the code-switching occurred at the word level. *Intra-word code-switching* (Myers-Scotton, 1989) performs code-switching within words. For example, in the words *'rajas', 'trophyian'*, the English morpheme *-s* appeared with the Hindi prefix *raja-* (English translation: king) and the Hindi morpheme *-ian* appears with English prefix *trophy*, both marking the plurality (Das & Gambäck, 2014).

### *2.4. Levenshtein distance*

Levenshtein distance is used to calculate the smallest estimate of operations done on the source string to mould it into the target string.

Formally, the Levenshtein distance between two strings, $s_1$ and $s_2$ (of length $|s_1|$ and $|s_2|$ respectively), is given by $\text{lev}_{s_1,\, s_2}(|s_1|,|s_2|)$ where:

$$\text{lev}_{s_1,s_2}(i,j) = \begin{cases} \max(i,j) & \text{if} \min(i,j) = 0 \\ \min \begin{cases} \text{lev}_{s_1,s_2}(i-1,j) + 1 \text{ (Deletion)} \\ \text{lev}_{s_1,s_2}(i,j-1) + 1 \text{ (Insertion)} \\ \text{lev}_{s_1,s_2}(i-1,j-1) + 1_{(s_{1i} \neq s_{2j})} \text{ (Substitution)} \end{cases} & \text{otherwise} \end{cases} \tag{1}$$

Here, the first case indicates that when there is no difference between characters of the two strings, i.e., $min(i,j)=0$, there is no increment to the count of edits. The second case is when the characters are different in the two strings, and the function accounts for a change or edit by incrementing 1 to the count of edits. We have used Levenstein distance to calculate the similarity between words in Section 5.3.2.

## 3. Related Work

### *3.1. Existing Hate Speech Detection Methods*

The researchers primarily cast hate speech detection as a supervised classification task (Schmidt & Wiegand, 2017). Numerous works perform document classification using classical machine learning algorithms, predominantly SVM, Naive Bayes, Logistic Regression and Random Forest (Burnap & Williams, 2015; Davidson et al., 2017; Mehdad & Tetreault, 2016; Waseem & Hovy, 2016). Also, there are many works citing performance improvements using Deep learning models, e.g. (uni- and bi-)LSTMs, GRUs, CNNs, and transformers (Gambäck & Sikdar, 2017; Park & Fung, 2017; Del Vigna12 et al., 2017).

*Classical Classification methods* show improvements when manually engineered features of data are fed into the models. Several state-of-the-art features have been covered extensively in research (Schmidt & Wiegand, 2017). We recapitulate the important features in Table 3.

*Deep Learning methods* have multi-layer structures with non-linearity that have shown improvement over classical models without the use of extensive feature engineering (Park & Fung, 2017; Del Vigna12 et al., 2017).

Table 3: Classical methods employing Feature Engineering

| Related Works | Features |
|---|---|
| (Davidson et al., 2017) | TF-IDF features, POS tags, Sentiment Score, Symbols and Characters count |
| (Burnap & Williams, 2016) | Bag of words, Char n-grams, Typed Dependencies, Hateful terms |
| (Zhong et al., 2016) | BOW, Sum of Grammatical dependency intensity level, Word2Vec |
| (Del Vigna12 et al., 2017) | Sentiment Polarity, Word Embedding |
| (Waseem & Hovy, 2016) | Text length, Gender or Geographic information, Character n-grams, User information |
| (Mehdad & Tetreault, 2016) | Lexical and Morphological Features |
| (Nobata et al., 2016) | N-grams, Linguistic, Syntactic and Distributional Semantics Features |

These models are fed either with the word or context embeddings pre-trained on deep neural networks (Mehdad & Tetreault, 2016; Yuan et al., 2016), or with simply the one-hot encoding as the input (Badjatiya et al., 2017; Zhang et al., 2018). Recurrent and Convolutional Neural Networks(RNN and CNN) are the most popular deep learning models adopted in the literature. CNN can perform word extractions and provides data parallelization (Gambäck & Sikdar, 2017), while RNN can model word dependencies (Del Vigna12 et al., 2017). Many works employ a hybrid of both these networks (Zhang et al., 2018; Zhang & Luo, 2019). More recently, the research focus has shifted to the new, more promising neural architecture called the Transformer (Vaswani et al., 2017). Many works employed Transformers due to the faster training speed compared to RNN and CNN and better interpretability of self-attention layers. Concerning hate speech detection, Bidirectional encoder representations from the transformers (Bert) model has been investigated on several English datasets (Mozafari et al., 2019). They introduced fine-tuning strategies, and the best performing model is Bert encoder layers fine-tuned with CNN. J. Zhu *et al.* used the neural network-based classifier fine-tuned on the pre-trained Bert model to detect offensive tweets (Zhu et al., 2019). It also tried offensive language detection with Bert. However, it reported their best results using BiLSTM with attention models. Another method synthesized labelled code-switched text (Samanta et al., 2019). The advent of transfer learning has motivated NLP research into pretraining language models on large corpora and then fine-tuning them for the specific task. Some recent works have made use of transfer learning for hate speech detection (Howard & Ruder, 2018b; Gröndahl et al., 2018).

*3.2. Existing works on Hindi-English code-switched text:*

There are three lines of work in the Hindi-English code-switching domain, and the following are their key insights as well as limitations:

1. Full-text translation to English: This is one of the techniques in which the Hindi words are first transliterated from Roman to Devanagari scripts, and then sentences are word-wise translated into English (Mathur et al., 2018). For example, *Tum ussey pyar kyu nahi karti?* is converted to *you him love why no?* This is a naive solution as it ignores syntax and grammatical rules and treats words in isolation.
2. Character level Translation: The Indic-Trans, (Bhat et al., 2015), is a transliteration library for multiple languages, including Roman Hindi to Devanagari Hindi and vice-versa. In this method, each letter of the word is individually translated from Roman to Devanagari script. While it provides a solution to tackle spelling variations for words, it is ineffective in transliterating Roman to Devanagari Hindi scripts due to the larger vocabulary size in the Devanagari script. This will result in incorrect transliterations. For example, 't' in Roman Hindi can be 'त' or 'ट' in Devanagari Hindi. This can produce prominent errors for words such as ‘tomato’, which can be written as ‘**tamatar**’ in Roman script, and character-level translation makes it तमातर in Devanagari script. On the contrary, the correct Devanagari Hindi word is टमाटर. We have experimented with character level translations in Section 6.

3. Feature Extraction: Previous works also attempt to identify hate in text using semantic (Badjatiya et al., 2017), syntactic and lexical based features such as *char n-grams*, *TF-IDF vectors*, *use of profane words*. However, models face constraints in detecting hate inducing texts, especially when the text does not contain profanity but inculcates hatred (Mathur et al., 2018). The second post of Table 2 is an example that insults the Bangladeshi community without profane words and hence is classified as hate speech.

### *3.3. Hate speech detection in Hindi-English code switched domain*

The publicly available Hindi-English hate speech datasets extracted from social media are the Hate Speech dataset (HS) (Bohra et al., 2018), the TRAC Hindi Facebook dataset (Kumar et al., 2018b), and the Hindi Offensive Tweet dataset (HOT) (Mathur et al., 2018). The first work on these datasets used the feature extraction technique and selected the character n-grams, punctuations, negation words and lexicon from the text and applied Support Vector Machine and Random Forest machine learning classifiers (Bohra et al., 2018). Mathur *et al.* performed full-text translation followed by Multi-Channel Transfer Learning model with word embeddings as well as features such as Sentiment Score and Linguistic Inquiry as input to their model (Mathur et al., 2018). Kamble *et al.* introduced domain-specific word embeddings (Kamble & Joshi, 2018), and reported their results on CNN, LSTM, and BiLSTM models. Prabhu *et al.* experimented with the sub-word level LSTM model on the Hinglish text (Prabhu et al., 2016).

## 4. Data collection and pre-processing

In this section, we provide the statistics of the three datasets which we have experimented with, also shown in Table 4, and the steps for their pre-processing.

### *4.1. TRAC-I*

Facebook is a social platform that has numerous aggressive Hindi and English posts. TRAC-I[1] (Kumar et al., 2018b) comprises of 12000 posts classified into Overtly Aggressive (this class overtly expresses aggression using typical symbolic lexicons), Covertly Aggressive (subtle, indirect expression of aggression including sarcasm, satire, and rhetorical questions) and Non-Aggressive. After pre-processing, the dataset was reduced to 11623 posts. TRAC-I has datasets formed from Facebook as well as Twitter. However, we used only the Facebook data as the other datasets that we have used are extracted from Twitter. This way, our experimentation is not specific to posts on one social media platform.

### *4.2. HS*

This data is collected from Twitter tweets (4575) and is labelled as hate speech(those tweets that induce hatred, irrespective of profanity) and normal speech (tweets that do not induce any form of hatred) (Bohra et al., 2018). This has Roman Hindi and English code-switched text. After text pre-processing, the number of tweets remained the same.

[1] https://sites.google.com/view/trac1/shared-task

Table 4: Pre-processed Datasets Description.

| Datasets | Labels Distribution | Posts from Datasets | Translation |
|---|---|---|---|
| HOT (Mathur et al., 2018) | Non-Offensive (1414) | haa jaise tum bhi abhi | Yeah as if you are |
| | Abusive (1942) | main jutt punjabi hoon aur imran league say nafrat karta hu | I am a strong Punjabi and hate Imran League (Muslims) |
| | Hate Inducing (323) | agar koi bole ki ja ke maar to s**le ki g**d maar lene ka | If someone says to kill then kill fool's a*s |
| HS (Bohra et al., 2018) | Hate Speech (1661) | main aise logo se nafrat karta hu jo caste ke sath jod ke chaude hote real vo piddu hote | I hate those people who associate with their caste and boast about it. In reality they are cowards |
| | Normal Speech (2914) | woh bhi samjhte hain hadh hoti hai nafrat ki | They also understand the extent of hate |
| TRAC-I (Kumar et al., 2018b) | Overtly Aggressive (4710) | gater ke keede | Gutter bugs |
| | Covertly Aggressive (4712) | yeh hai ache din | Are these the good days? |
| | Non-Aggressive (2212) | ab ye kaun h | Now who is this? |

### *4.3. HOT*

The HOT dataset (Mathur et al., 2018) consists of 3679 tweets classified into three categories which are Non-Offensive (a general form of speech that does not offend anybody), Abusive (offensive speech to hurt the sentiments of the receiver) and Hate Inducing (tweets that incite hate without profanity). Like the HS dataset, this has Roman Hindi and English code-switched text. After text pre-processing, the HOT dataset reduced to 2803 tweets.

Since we are focusing on tackling hate in text, we removed all nontextual information from the dataset. Data cleaning (Figure 1) involved the following steps:

1. We lower-cased and removed empty rows from all datasets.

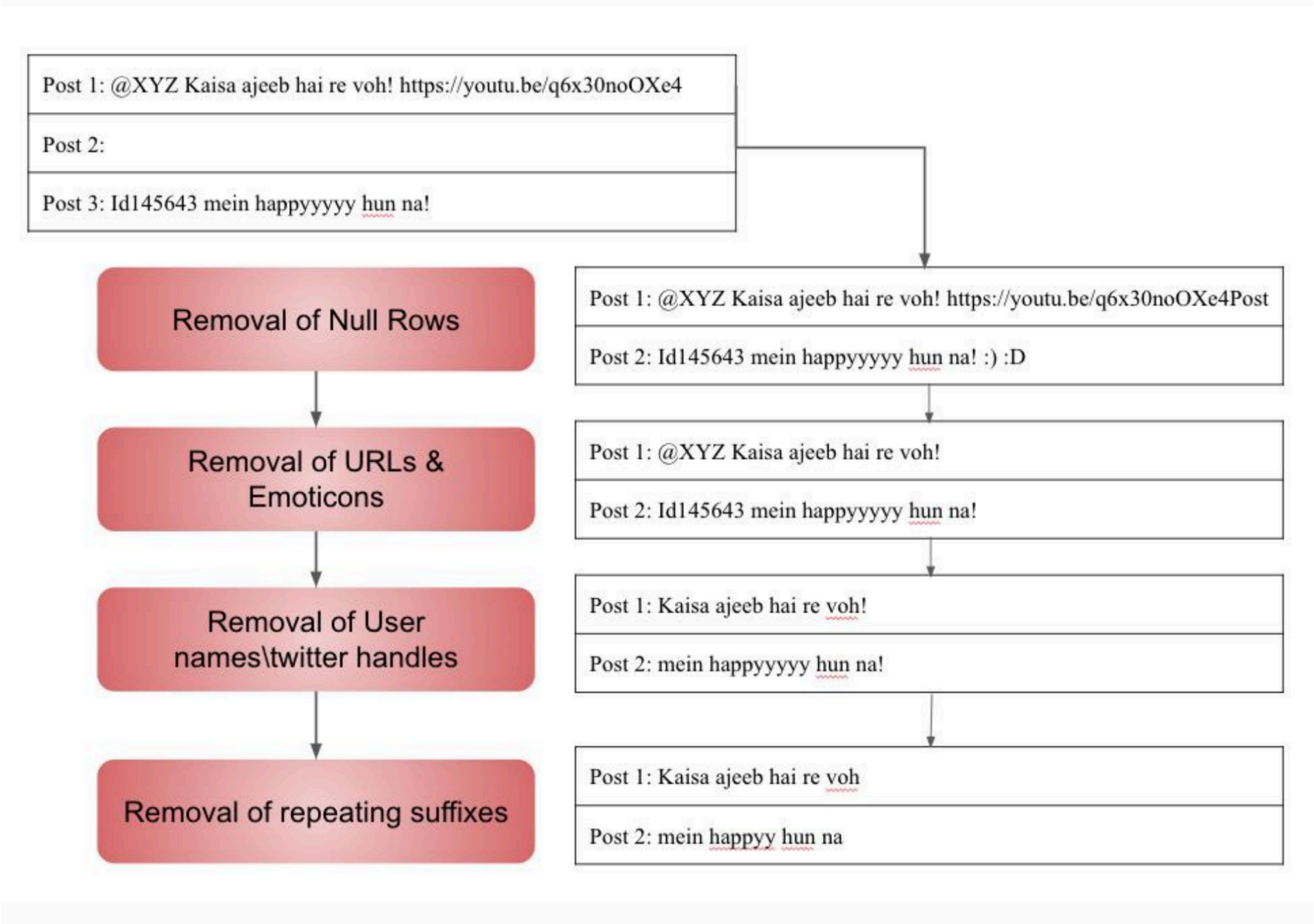

Figure 1: Steps of Pre-Text Processing

2. Removal of URLs: Since the datasets were extracted from social media sites, they consisted of tweets/posts where users shared URLs and tagged others. For example: '@..(name of individual)...URL'. Keeping a check on what social media users share through hyperlinks is kept out of this paper's scope.

3. Removal of name/user_id was done as it did not contribute to that post's content. Similarly, we also discarded the non-textual elements such as emoticons and symbols, because they can be misleading (Griessel, 2021)

4. Many social media users use multiple alphabets to express extreme emotions such as: 'Today I am so happyyyy...' rather than, 'Today I am so happy...'. So we normalized these words to set the duplicate consecutive characters order to two only. This way we retained the user's implicit emotions while ensuring that there are no multiple word representations for the same words (Cui et al., 2011). Therefore words in tweets with multiple characters such as happyyy, are changed to happyy.

## 5. Proposed Work

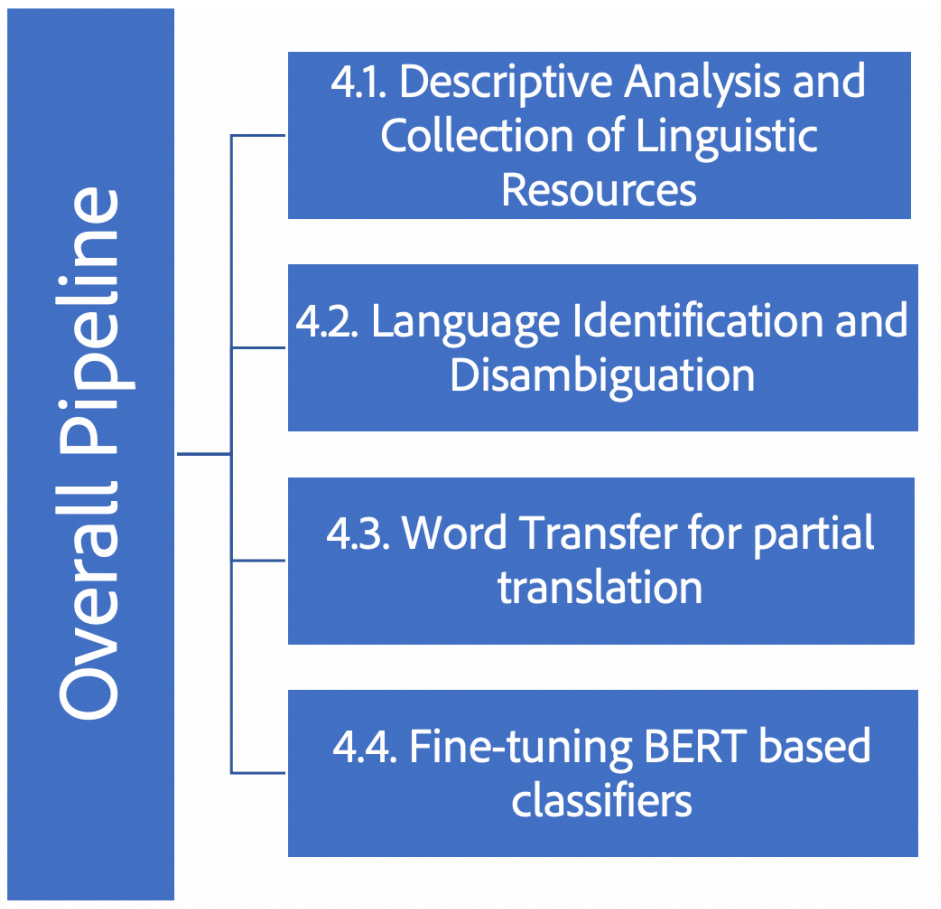

Figure 2: Overall Pipeline of *MoH* + Bert based models

We have divided our proposed work into four significant steps as shown in Figure 2. Each step is clearly illustrated in this section.

### *5.1. Descriptive Analysis*

We began with statistical analysis to study the text structure in all datasets: TRAC-I, HS, and HOT. Since the language mixing occurs in individual tokens, we identified the datasets as code-switched rather than other language forms such as borrowing, pidgins, creoles, or loan translation (Gumperz, 1982; Poplack & Sankoff, 1984; Muysken, 1995).

1. The first observation from the raw dataset is that some posts have grammatical errors and even unstructured sentences. For example, the following post *'mumbai kahan duniya'* translates to, 'Mumbai where world'. This sentence brings ambiguity because of the sentence's improper structure and the lack of essential parts of speech. It is imperative to work with well-formed sentences in order to capture covert acts of hate.
2. The second observation focused on the types of code-mixing (Refer Section 2.3). The posts that alternatively contain English and Hindi sentences employ inter-sentential code-switching. For example, *'iss ko muj se kya problem, khud to kuch ata nahi, meri performance ko bekar bolta hai'* which means, 'What is his problem with me? He knows nothing, yet criticizes my performance.' Here, the user inserted English words (*'problem'* and *'performance'*) into the grammatical structure of Roman Hindi. A few posts made use of *intra-word code-switching.* For example, *'never donate to any priests they are chors'*. The English morpheme *-s* appeared with the Hindi prefix *chor-* (English translation: thief), which marks plurality.

### *5.2. Language Identification and Disambiguation*

Language identification is an essential step in a multilingual task. Due to the text's code-switched nature, we used word-based language tracking to

[2] Refer Algorithm 2 for generation of *MoH* Knowledge Base

**Algorithm 1:** *MoH* Language Identification and Disambiguation

```
Input: English, Roman Hindi and Devanagari Hindi code-switched text.
Output: correctly spelled English and Devanagari Hindi code-switched text.
Method:
for each word, 'w' in input text do
    if Enchant Hindi detects w then
        Assign label 'DEV HINDI' to w
    end
    if Enchant Hindi spell checks w then
        Assign the correct spelling to w
        Assign label 'DEV HINDI' to w
    end
    if Enchant English detects w then
        w_eng = w
        isEnglish = True
    end
    if Enchant English spell checks w then
        w_eng = w
        Assign correct spelling to w
        isEnglish = True
    end
    if w in MoH Knowledge Base[2] then
        w_hin = MoH Knowledge Base[w]
        isHindi = True
    end
    if isEnglish == True and isHindi == False then
        w = w_eng
        Assign label 'ENGLISH' to w
    end
    if isEnglish == False and isHindi == True then
        w = w_hin
        Assign label 'ROM HINDI' to w
    end
    if isEnglish == True and isHindi == True then
        freq_eng = n_gramFrequency(w_eng)
        freq_hin = n_gramFrequency(w_hin)
        if freq_eng > freq_hin then
            w = w_eng
            Assign label 'ENGLISH' to w
        else
            Assign w_hin to w : w = w_hin
            Assign label 'ROM HINDI' to w
        end
    end if isEnglish == False and isHindi == False then
        Assign label 'OOV' to w
    end
end
```

differentiate between Roman Hindi and English words. Previous works have employed conversational features (Sarma et al., 2019), hybrid-based techniques (Banerjee et al., 2014) and corpus-based techniques (Beatty-Martínez et al., 2018) for language identification.

*5.2.1. Using Enchant*

Our method makes use of the Python Library Enchant (Kelly, 2016) for language identification. Enchant also suggests spelling changes for slightly misspelt words and hence acts like a spell checker. However, Enchant only works with text written in the standard language script. That means if Enchant has to detect Hindi words, they should be written in the Devanagari script. Since our data consists of Roman Hindi words, we created a comprehensive knowledge base (Section 5.2.2) for language identification of these words.

*5.2.2. Collection and Summarization of Linguistic Resources*

We used two existing knowledge bases for creating an exhaustive Roman Hindi vocabulary and call it the *MoH* knowledge base[3]:

1. FIRE Track public transliteration pairs (of size 30,823), which is a Hindi words list in Roman, and Devanagari Scripts (Roy et al., 2013). These pairs were derived using the alignment of Hindi song lyrics.
2. XLIT-Crowd transliteration pairs of IIT Bombay (Kunchukuttan et al., 2017). This data was extracted via crowdsourcing on Amazon Mechanical Turk. The Roman-Devanagari Hindi pairs were of size 14,919 pairs. This corpus also included Devanagari Hindi sentences and their respective English translations. To keep a consistent knowledge base repository, we normalized all the sentences into tokens and added these tokens to the existing parallel words dictionary.

[3]https://github.com/anubhakabra/MoH_Hate_Speech_Detection.git

3. Although these resources have vastly covered most of the Hindi language tokens, we observed that they did not include Hindi profane words on inspection. Since the datasets consist of profane words, we included the Hindi profane words list into the knowledge bases (Mathur et al., 2018).

We observed that some of the knowledge base words are used only in the English language, e.g., *empowered*. We iterated on these pairs and checked if the key is detected by Enchant English Detector and not by detected Enchant Hindi Detector. If so, the key-value pair was removed from the knowledge base. Finally, we have a knowledge base of 72,635 transliteration pairs, where the keys are the Hindi words in Roman script, and values are the transliterated words in Devanagari script. This section is summarized in Algorithm 2.

**Algorithm 2:** Generation of *MoH* Knowledge Base

*Given:* Firetask (F), XLIT-Crowd (XLIT), and Profane Words (PW)

*MoH* Knowledge Base = F ∪ XLIT ∪ PW ;

**for** *English sentences and Hindi translations in MoH Knowledge Base* **do**

  Split into words;

  MoH Knowledge Base[Eng_word] = Hin_translation;

**end**

**for** *word and its Hindi translation in MoH Knowledge Base* **do**

  **if** *Enchant English detects word and Enchant Hindi doesn't detect Hindi translation* **then**

    *MoH* Knowledge Base - {word and Hindi translation};

**end**

*5.2.3. Word Disambiguation*

Word disambiguation is essentially labelling each word with its respective language tags without any ambiguities. Figure 3 and Algorithm 1 shows the Word Disambiguation process in detail.

1. In the first case of token categorization, if the word was detected by Enchant English or Hindi detector, it was assigned the 'ENGLISH' or 'DEV HINDI' language tags. If the word was identified only by our knowledge base, it was given the 'ROM HINDI' tag. If the word remained undetected, it was labeled 'OOV', i.e., the out of vocabulary tag. The OOV tagged words are Roman Hindi words that are spelt differently, compared to the Roman Hindi spelling in *MoH* Knowledge Base, or they are misspelt English words.
2. The second case is that the word existed in *MoH* Knowledge Base and was also detected by Enchant English detector, i.e., it could ei-

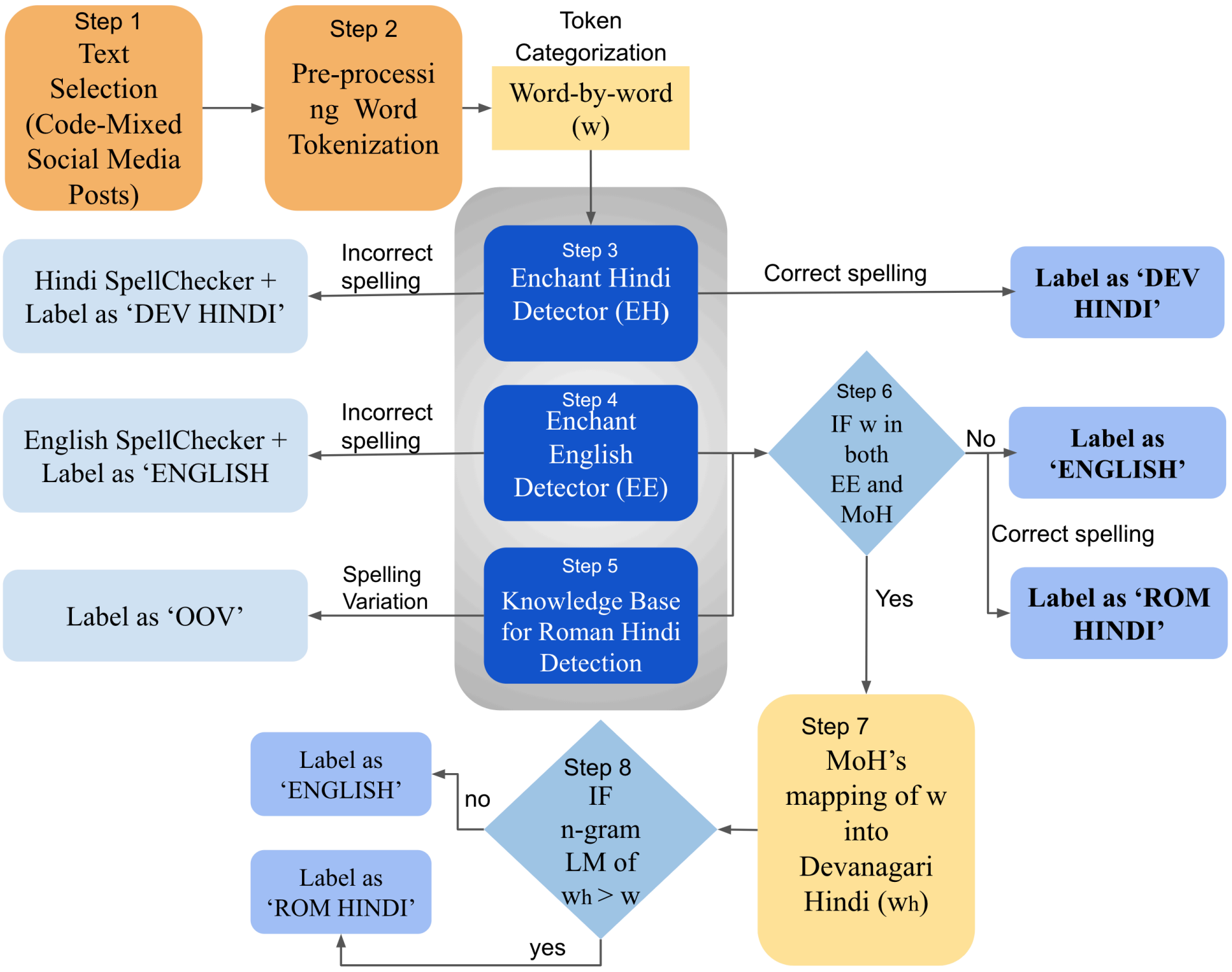

Figure 3: Language Identification and Disambiguation

ther be an English or a Hindi word written in Roman script. For example, 'tum' means 'you' in Hindi and 'abdomen' in English. We took these words and their corresponding Devanagari translations from *MoH* Knowledge Base. We then checked the frequency of these words in the n-gram English and Hindi language models for word disambiguation. Whichever words showed a higher frequency of use, we selected their respective languages as the language tags.

English and Devanagari Hindi word spellings were checked during the identification procedure, and all words were labelled with their respective language tags. This defined how each word was treated for further processing. There is no spelling error corrector for words written in Roman

**Algorithm 3:** Transfer of OOV words.

```
Input : The words labelled as 'OOV'
Output : Label identification of these words
for word (w) labelled as OOV do
    max_sim=0
    right_word = None
    for candidate word (c) in MoH_knowlege_base do
        sim = Lev(w,c)
        if sim > 0.70 and sim>max_sim then
            max_sim = sim
            right_word = c
        end
    end
    if right_word is not None then
        Replace w with right_word
        Assign label 'ROM HINDI' to w
        Transfer MoH[w] in place of w
    else
        Assign label 'NA' to w
    end
end
```

To optimize the time complexity of Algorithm 3, we vectorized these operations using NumPy (frompyfunc) and Multiprocessing libraries in Python 3.6

Hindi, simply because there is no standard Roman Hindi script. Users generally tend to write different spellings according to their speaking styles. This is commonly seen with the use of vowels. Nevertheless, we have performed steps to correctly identify the Roman Hindi words currently marked as 'OOV', mentioned in Section 5.3.2.

### *5.3. Word transfer for partial translation*

On inspection of words labelled as 'OOV', we found that most of these were present in *MoH* knowledge base but were not identified due to the spelling difference. To curb this, we used Levenshtein Distance Similarity. If the similarity score of the 'OOV' tagged words with the *MoH* knowledge base words was higher than the threshold (0.70), these words were labelled as 'ROM HIN', else they were assigned 'NA' tag. Refer Algorithm 3.

*5.3.1. Threshold Assignment*

Since most of the OOV words are differently spelt Roman Hindi text, we analyzed these words before choosing the threshold for the text similarity algorithm.

The spelling differences in Roman Hindi arise due to the insertion, substitution or deletion of vowels in text. For example, 'm**au**sam' (weather) can be written as 'm**o**sam' but not as '**r**ausom'. The '*au*' and '*o*' sounds are similar in Hindi and hence are interchanged in informal Roman Hindi script. Generally, the difference between the source and target words pertain to a few positions of vowels. Hence we experiment with a Levenshtein similarity score above 0.5. We sampled 20% of total posts randomly from all datasets and passed them through the MoH pipeline. After which we conducted a manual inspection to see the percentage of correctly Roman to Devanagari Hindi mapped words: a threshold value of 0.6 and 0.8 correctly translated 76% and 68% of words, respectively; however, 0.7 value captured 91% of the words.

On working with values lesser than 0.7, we observed that unrelated but similar Devanagari Hindi words replaced many proper nouns such as names of people or places. For example, '*tom*' (proper noun) was being replaced by '*tum*' (you), which had a similarity of 67%. For a similarity score above 0.7, important words were getting missed, e.g. *batey* and *batein* are both acceptable Roman Hindi spellings of the English word 'conversations' but have similarity of 72%. The experimentation on the three datasets showed that 0.7 worked considerably well for Hindi-English code switched tasks. However, this parameter can be tuned according to the task or language.

#### *5.3.2. Correlating spellings using Levenshtein Distance*

A standard norm in spelling checkers is to use edit distance to detect how dissimilar the two strings are. Different variants of edit distance tackle different string operations. We used one of these variants, 'Levenshtein Distance' (Refer Section 2.4 for more details). Other distance variants such as Longest Common Subsequence (Insertion and Deletion operations only) and Hamming Distance (Substitution operation only), did not perform well in our case and hence were not used. Table 5 shows the Levenshtein conversions of the source words (Roman Hindi words mapped as 'OOV' taken from the datasets) into the target words (data in *MoH* knowledge base). The first three source words are different Roman Hindi word namaste spellings, which translates to 'Hello' in English. These have high similarity scores with the 'namaste' word present in *MoH*, while the last source word **nafrat**, which means 'hate' in English, has less similarity score to the 'namaste' target word. Hence, 'nafrat' is not converted into namaste.

Table 5: 'Namaste' is a greeting in Roman Hindi, which means 'Hello' in English. Here, the source words are different writing styles, the same word as found in datasets, and the target words are the words present in *MoH* Knowledge Base. LD refers to Levenshtein Distance between the two words.

| Source Word | Target word | Operation | LD | Similarity | OOV Data Conversion |
|---|---|---|---|---|---|
| namste | namaste | Addition | 1 | 86% | Yes |
| namastey | namaste | Deletion | 1 | 88% | Yes |
| namuste | namaste | Substitution | 1 | 86% | Yes |
| nafrat | namaste | None | 4 | 43% | No |

Subsequently, *MoH* knowledge base is used to replace all 'ROM HINDI' tagged words into their respective Devanagari Hindi translations. This becomes possible as for every 'ROM HINDI' tagged word, now there is a corresponding Roman to Devanagari Hindi mapping in the *MoH* knowledge base.

### *5.4. Hate speech classification using Bert*

Natural language processing has shifted its attention to transfer learning for a boost in performance and decrease in training time (Howard & Ruder, 2018a; Ruder et al., 2019). Transfer learning is key to a new era of deep learning solutions in natural language processing because it solves the bottlenecks caused by a shortage of labelled training data.

Pre-training alleviates the load of understanding the language semantics from limited data. Language models are already pre-trained on a vast corpus, and so they can directly fine-tune a smaller labelled dataset for the task in hand. Since transfer learning's improvements are derived from the models' pre-training, it is essential to understand that the language model transfers knowledge to a task in the same language as that of the pre-training language model (Pires et al., 2019).

1. M-Bert (Multilingual Bert) is a pre-trained language model trained on the Wikipedia corpus of 104 languages (Pires et al., 2019).

   This model's key achievement is that even though it was pre-trained on distinct 104 monolingual corpora, it can transfer to languages with no shared vocabularies, such as English and Devanagari Hindi. Moreover, it is empirically shown that M-Bert transfers well on code-switched text compared to the Roman Hindi mixed with English text because the model has developed more profound multilingual representation (Pires et al., 2019).

2. MuRIL (Multilingual Representations for Indian Languages) is a newly introduced Bert model pre-trained on 17 Indian languages, including Hindi and English, and their transliterated counterparts. It is a modification of M-Bert for low-resource Indian languages, as it includes pairs of translated and transliterated texts while training.

According to the distributional hypothesis, if the same word regularly occurs in different contexts, we may assume the word is polysemous (Miller & Charles, 1991). Static word embeddings are made by iteratively learning embeddings, such that each word gets a single representation. This becomes an issue as a single word embedding vector will need to represent all the polysemous words. Hence word sense disambiguation step becomes crucial in static embeddings (Ethayarajh, 2019). The advantage of Bert based models is that they can perform word sense disambiguation due to the contextual word embeddings (Hadiwinoto et al., 2019; Zhu, 2020).

#### *5.4.1. Model Training*

We first tokenized the *MoH* output data using Bert's Word Piece tokenizer to split words into subwords. We kept the maximum sequence length as 128 and skipped the tweets whose length exceeded the maximum length due to memory constraints. We converted the input representation for each tweet into its Bert embedding, which we then passed through a combination of dropout and dense layers. The model was implemented in Keras and TensorFlow (TensorFlow, 2018) and optimized using the AdamW optimizer. After training, we computed the performance using evaluation metrics: F1 score, Precision and Recall.

In the experimentation, we tune our hyperparameters such that the Batch Size is 32, Learning Rate is 2e-5, Dropout Rate is 0.2, Dropout Layers are 2, Dense Layers are 3, Number of epochs are 10, warmup is 0.1, Optimizer is AdamW and Activation functions are Relu and Softmax. We used binary crossentropy loss for datasets with two classes, and categorical crossentropy loss for multi-class datasets.

## 6. Experimentation and Results

The following section describes the experiments that will help understand how the *MoH* mapping technique and the Bert based models individually contributed to the best results. First, we compared performance with classical machine learning models where we used simple surface features. We tested these models on the raw data and then on proposed *MoH* mapped data. Second, we compared the proposed work with baseline models. Third, we compared other data transliteration techniques with the proposed technique. We trained these data transliterations on the same Bert based models that we used to train *MoH* mapped data. For ease of understanding, we have used the abbreviation *MoH + M-Bert* and *MoH + MuRIL* to represent our techniques. The results and findings are elaborated below:

### *6.1. Simple Surface features with ML classifiers*

We utilized frequency-based and count-based surface-level features that have been used extensively in hate speech classification tasks (Burnap & Williams, 2015; Van Hee et al., 2015; Waseem & Hovy, 2016; Burnap & Williams, 2016; Nobata et al., 2016).

These features are simple and generally very good at predictive modelling (Schmidt & Wiegand, 2017).

First, we performed feature engineering on the datasets to extract Count vectors, TF-IDF Word level vectors, Word level n-gram vectors, and Character level vectors. These are the most popular surface features (Schmidt & Wiegand, 2017). Then using individual features, we performed classification with state-of-the-art Machine Learning models: Naive Bayes, Logistic Regression, SVM, Random Forest, XGBoost, AdaBoost, and Gradient Boosted Decision Trees. The best values of Precision, Recall, and F1 are bolded in Table 6 and a comparative study between features is shown in Figure 4.

The features we have used vary from word counts and frequencies to character counts and frequencies. From Table 6, we infer that character level vectors on SVM classifier for TRAC-I data, count vectors on Logistic Regression for HOT data, and n-gram vectors on Logistic Regression for HS data give the best F1 score. This implies that the features and the classifiers perform differently on different datasets. Hence, we cannot adopt a single feature on a particular machine learning-based classifier for this task.

We then used the datasets refined using *MoH* mapping technique (steps mentioned in Section 5). *MoH* converted Roman Hindi into Devanagari Hindi script, so the final data's vocabulary consisted of either Devanagari Hindi or English words. We selected the same features and performed classification using all of the classifiers mentioned above. The trend of scores is similar to that of the previous experiment, as shown in Table 7. However, it is interesting to observe that scores are slightly higher with *MoH* mapped datasets.

Table 6: Experimentation on TRAC-I, HOT, and HS Datasets *(without running through MoH pipeline)*. Showing Surface Features with Classical ML models. P: Precision, R: Recall, F: F1 Score. Bolded values show the highest value for that particular metric.

| | | **TRAC-I Data** | | | **HOT Data** | | | **HS Data** | | |
|---|---|---|---|---|---|---|---|---|---|---|
| **Model** | **Feature** | **P** | **R** | **F** | **P** | **R** | **F** | **P** | **R** | **F** |
| Naive Bayes | Count Vectors | 0.58 | 0.53 | 0.62 | 0.70 | 0.65 | 0.62 | 0.63 | 0.62 | 0.51 |
| | WordLevel TF-IDF | 0.52 | 0.50 | 0.49 | 0.48 | 0.49 | 0.48 | 0.56 | 0.67 | 0.54 |
| | N-Gram Vectors | 0.43 | 0.59 | 0.45 | 0.47 | 0.80 | 0.49 | **0.64** | 0.70 | 0.64 |
| | CharLevel Vectors | 0.51 | 0.59 | 0.51 | 0.49 | 0.48 | 0.48 | 0.55 | 0.71 | 0.51 |
| Logistic Regression | Count Vectors | 0.56 | 0.56 | **0.75** | **0.83** | **0.78** | **0.64** | 0.65 | **0.64** | 0.55 |
| | WordLevel TF-IDF | 0.54 | 0.59 | 0.55 | 0.61 | **0.83** | 0.64 | 0.62 | 0.69 | 0.63 |
| | N-Gram Vectors | 0.45 | 0.56 | 0.47 | 0.40 | 0.50 | 0.38 | 0.62 | **0.75** | 0.62 |
| | CharLevel Vectors | **0.57** | 0.59 | **0.57** | 0.60 | 0.78 | 0.62 | 0.61 | 0.66 | 0.61 |
| SVM | Count Vectors | 0.46 | 0.56 | 0.48 | 0.33 | 0.21 | 0.26 | 0.5 | 0.32 | 0.39 |
| | WordLevel TFIDF | 0.55 | 0.59 | 0.56 | 0.33 | 0.21 | 0.26 | 0.5 | 0.32 | 0.39 |
| | N-Gram Vectors | 0.44 | **0.60** | 0.46 | 0.33 | 0.21 | 0.26 | 0.5 | 0.32 | 0.39 |
| | CharLevel Vectors | **0.57** | 0.55 | 0.56 | 0.33 | 0.21 | 0.26 | 0.5 | 0.32 | 0.39 |
| Random Forest | Count Vectors | 0.54 | 0.57 | 0.55 | 0.65 | 0.77 | 0.68 | 0.62 | 0.70 | 0.62 |
| | WordLevel TF-IDF | 0.52 | 0.53 | 0.51 | 0.65 | 0.77 | 0.69 | 0.61 | 0.63 | 0.64 |
| | N Gram Vectors | 0.48 | 0.53 | 0.49 | 0.49 | 0.56 | 0.50 | 0.62 | 0.63 | 0.59 |
| | Char Level Vectors | 0.49 | 0.57 | 0.50 | 0.63 | 0.70 | 0.67 | 0.53 | 0.65 | 0.58 |
| XGBoost | Count Vector | 0.53 | 0.53 | 0.52 | 0.59 | 0.67 | 0.65 | 0.59 | 0.68 | 0.59 |
| | WordLevel TF-IDF | 0.47 | 0.59 | 0.46 | 0.68 | 0.76 | 0.65 | 0.59 | 0.65 | 0.61 |
| | N Gram Vectors | 0.38 | 0.59 | 0.43 | 0.35 | 0.46 | 0.36 | 0.60 | 0.72 | 0.62 |
| | CharLevel Vectors | 0.52 | 0.54 | 0.52 | 0.71 | 0.77 | 0.75 | 0.61 | 0.66 | 0.62 |
| GradientBoosting | Count Vectors | 0.51 | 0.50 | 0.56 | 0.67 | 0.61 | 0.60 | 0.66 | 0.58 | 0.50 |
| | WordLevelTF-IDF | 0.51 | 0.57 | 0.48 | 0.58 | 0.70 | 0.58 | 0.59 | 0.70 | 0.61 |
| | N-Gram Vectors | 0.40 | 0.58 | 0.37 | 0.38 | 0.62 | 0.38 | 0.60 | 0.70 | 0.64 |
| | CharLevelVectors | 0.55 | **0.60** | 0.54 | 0.63 | 0.68 | 0.64 | 0.59 | 0.64 | 0.58 |
| AdaBoost | Count Vectors | 0.57 | 0.50 | 0.54 | 0.62 | 0.51 | 0.55 | 0.57 | 0.58 | 0.49 |
| | WordLevel TFIDF | 0.48 | 0.57 | 0.47 | 0.56 | 0.64 | 0.59 | 0.55 | 0.62 | 0.60 |
| | N-Gram Vectors | 0.40 | 0.56 | 0.35 | 0.40 | 0.58 | 0.40 | 0.62 | 0.67 | 0.60 |
| | CharLevel Vectors | 0.50 | 0.53 | 0.52 | 0.65 | 0.64 | 0.62 | 0.55 | 0.54 | 0.59 |
| *MoH* + M-Bert | - | **0.74** | **0.60** | **0.66** | **0.84** | **0.93** | **0.88** | **0.83** | **0.80** | **0.81** |
| *MoH* + MuRIL | - | **0.79** | **0.67** | **0.71** | **0.87** | **0.95** | **0.90** | **0.86** | **0.84** | **0.85** |

Table 7: Experimentation on TRAC-I, HOT, and HS Datasets which are transformed using ***MoH* Knowledge Base**. Showing Surface Features with Classical ML models. P: Precision, R: Recall, F: F1 Score. Bolded values show the highest value for that particular metric.

| | | **TRAC-I Data** | | | **HOT Data** | | | **HS Data** | | |
|---|---|---|---|---|---|---|---|---|---|---|
| **Model** | **Feature** | **P** | **R** | **F** | **P** | **R** | **F** | **P** | **R** | **F** |
| Naive Bayes | Count Vectors | 0.58 | 0.54 | 0.63 | 0.75 | **0.67** | 0.67 | 0.64 | 0.65 | 0.54 |
| | WordLevel TF-IDF | 0.54 | 0.52 | 0.51 | 0.51 | 0.50 | 0.50 | 0.58 | 0.70 | 0.57 |
| | N-Gram Vectors | 0.47 | 0.60 | 0.47 | 0.49 | **0.85** | 0.51 | 0.66 | 0.72 | 0.66 |
| | CharLevel Vectors | 0.54 | 0.59 | 0.53 | 0.53 | 0.50 | 0.51 | 0.57 | 0.73 | 0.58 |
| Logistic Regression | Count Vectors | 0.58 | 0.58 | **0.75** | **0.85** | **0.80** | **0.67** | 0.65 | 0.65 | 0.57 |
| | WordLevel TF-IDF | 0.54 | 0.60 | 0.57 | 0.62 | 0.84 | 0.66 | 0.63 | 0.71 | 0.65 |
| | N-Gram Vectors | 0.47 | 0.58 | 0.49 | 0.42 | 0.50 | 0.40 | 0.61 | **0.78** | 0.65 |
| | CharLevel Vectors | **0.59** | 0.60 | **0.59** | 0.62 | 0.78 | 0.64 | 0.62 | 0.68 | 0.63 |
| SVM | Count Vectors | 0.48 | 0.56 | 0.50 | 0.35 | 0.23 | 0.28 | 0.57 | 0.30 | 0.43 |
| | WordLevel TFIDF | 0.57 | 0.59 | 0.58 | 0.35 | 0.26 | 0.28 | 0.52 | 0.35 | 0.41 |
| | N-Gram Vectors | 0.46 | **0.61** | 0.47 | 0.35 | 0.23 | 0.28 | 0.5 | 0.33 | 0.41 |
| | CharLevel Vectors | **0.59** | 0.56 | 0.57 | 0.38 | 0.27 | 0.31 | 0.49 | 0.38 | 0.39 |
| Random Forest | Count Vectors | 0.58 | 0.58 | 0.57 | 0.67 | 0.75 | 0.69 | 0.63 | 0.72 | 0.65 |
| | WordLevel TF-IDF | 0.53 | 0.54 | 0.53 | 0.67 | 0.78 | 0.70 | **0.67** | 0.68 | **0.67** |
| | N Gram Vectors | 0.45 | 0.55 | 0.46 | 0.50 | 0.57 | 0.52 | 0.63 | 0.65 | 0.61 |
| | Char Level Vectors | 0.51 | 0.58 | 0.52 | 0.64 | 0.71 | 0.68 | 0.55 | 0.66 | 0.58 |
| XGBoost | Count Vector | 0.56 | 0.55 | 0.55 | 0.59 | 0.68 | 0.66 | 0.60 | 0.68 | 0.60 |
| | WordLevel TF-IDF | 0.49 | 0.60 | 0.48 | 0.69 | 0.78 | 0.67 | 0.61 | 0.66 | 0.63 |
| | N Gram Vectors | 0.39 | 0.59 | 0.43 | 0.37 | 0.48 | 0.42 | 0.61 | 0.70 | 0.62 |
| | CharLevel Vectors | 0.53 | 0.55 | 0.53 | 0.72 | 0.78 | 0.77 | 0.63 | 0.68 | 0.64 |
| GradientBoosting | Count Vectors | 0.53 | 0.54 | 0.57 | 0.68 | 0.63 | 0.62 | 0.67 | 0.60 | 0.52 |
| | WordLevelTF-IDF | 0.53 | 0.58 | 0.49 | 0.60 | 0.72 | 0.60 | 0.60 | 0.71 | 0.62 |
| | N-Gram Vectors | 0.43 | 0.59 | 0.39 | 0.39 | 0.64 | 0.39 | 0.61 | 0.72 | 0.65 |
| | CharLevelVectors | 0.56 | **0.61** | 0.55 | 0.64 | 0.69 | 0.66 | 0.61 | 0.67 | 0.60 |
| AdaBoost | Count Vectors | 0.56 | 0.51 | 0.55 | 0.64 | 0.52 | 0.56 | 0.58 | 0.59 | 0.51 |
| | WordLevel TFIDF | 0.44 | 0.58 | 0.49 | 0.57 | 0.66 | 0.61 | 0.57 | 0.63 | 0.61 |
| | N-Gram Vectors | 0.42 | 0.57 | 0.37 | 0.40 | 0.59 | 0.42 | 0.63 | 0.68 | 0.62 |
| | CharLevel Vectors | 0.52 | 0.55 | 0.55 | 0.66 | 0.66 | 0.65 | 0.54 | 0.58 | 0.61 |
| *MoH* + M-Bert | - | **0.74** | **0.60** | **0.66** | **0.84** | **0.93** | **0.88** | **0.83** | **0.80** | **0.81** |
| *MoH* + MuRIL | - | **0.79** | **0.67** | **0.71** | **0.87** | **0.95** | **0.90** | **0.86** | **0.84** | **0.85** |

Table 8: Comparison between Original Data and *MoH* mapped Data features. Here the same Hindi word 'लिए' had two different spellings in original Roman Hindi data: 'liye' and 'lye', so the count vectors are different for the same word in original data.

| | | | |
|---|---|---|---|
| Word 1: | iss **lye** | *MoH* translation: | इस लिए |
| Count Vector: | [0,...,2932,...,1242,...0] | Count Vector: | [0,...,3782,...,2853,...0] |
| Word 2: | iss **liye** | *MoH* translation: | इस लिए |
| Count Vector: | [0,...,2932,...,3772,...0] | Count Vector: | [0,...,3782,...,2853,...0] |

To understand this phenomenon, we compared the word representations of the original dataset, and *MoH* mapped dataset. Table 8 shows that the count vectorizer assigns different count to the same words that are spelt differently (लिए spelt as **lye** or **liye**) in the original data, thus viewing those as new words and assigning a different count to them. On the other hand, *MoH* maps the multiple Roman spellings of the same words into the same Devanagari spelling. So the count vectorizer gives the same word representations to these words that had earlier different vectors. This means that the different Roman spellings for the same Hindi words are a limitation, and converting Roman Hindi to Devanagari Hindi is crucial in tackling code-switched text (Mehmood et al., 2020; Medrouk & Pappa, 2017).

Figure 4 shows the clear difference between the best performing surface features scores and the proposed, *MoH* + Bert based model scores. We see that the Precision, Recall and F1 values increase significantly for Bert based techniques for all the datasets. With *MoH* mapped data, there is an average increase of 13% between best surface feature-based scores and Bert based scores, and an increase of 4% when compared to using *raw data* with all the machine learning models.

As shown in Table 8 and Figure 4, it is clear that after refining data using *MoH*, the features extracted are much more refined and hence improved training of the classification models.

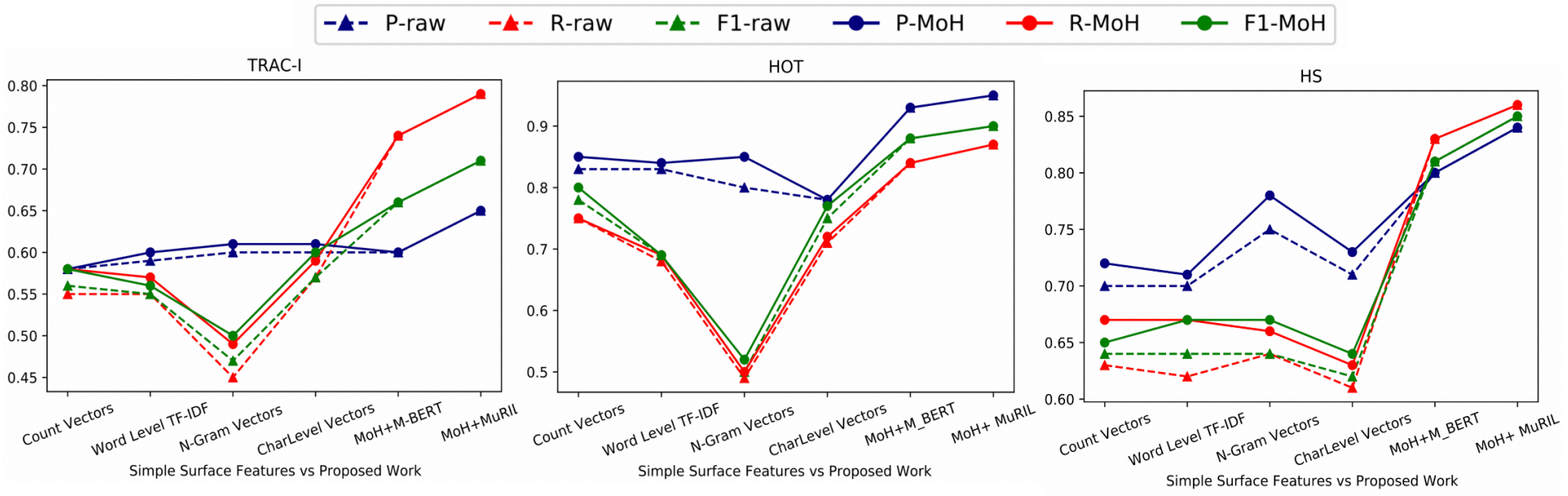

Figure 4: Comparison of simple surface features' performance with each other and proposed techniques. The best Precision(P), Recall(R) and F1 values are chosen for each feature irrespective of the ML model. Here, **raw** represents data which was NOT transformed by MoH. **MoH** represents data transformed by MoH before classification.

It can be inferred from both Table 6 and 7 that our technique supersedes in performance. Surface features do not capture semantics and sentence meaning. This is considered in Bert-based models, and hence they outperform classification results on extracted features.

### *6.2. Comparing with existing baselines*

We compared our performance with previous works on the three datasets for the next set of experiments: TRAC-I, HS, and HOT. We see that in Table 9, MoH + MuRIL has outperformed all the baselines for all the metrics. The prominent state-of-the-art models for each dataset are shown below:

#### *6.2.1. TRAC-I Baselines*

1. M-Bert (Modha et al., 2020): Their best results were obtained using M-Bert on raw data for classification.
2. n-gram + SVM (Kumar et al., 2018a): They used an SVM model trained on character and word bag-of-n-grams. Overlapping features use sublinear tf-idf to be weighed before usage.

3. Dense NN (Raiyani et al., 2018): This work proposed a technique that uses dictionary index and then represented the index values using one-hot encoding. The data was then passed into a dense neural network.
4. FeatureVec + LR (Samghabadi et al., 2018): They trained a Logistic Regression classifier on feature vectors such as unigrams, tf-idf, n-grams, and Google news pertained word embedding model. They transliterated Devanagari Hindi text to Roman Hindi.
5. Feature extraction + CNN (Singh et al., 2018): They extracted text, numerical and aggression based features and used these as input to a CNN model.
6. fastText + CNN (Modha et al., 2018): They used fastText embeddings on raw data and passed embeddings through Convolutional Neural Network layers.

*6.2.2. HS Baselines*

1. Synthetic + Gold (Samanta et al., 2019): This method generated synthetic text for code-switched data. They did so by generating automatic labelled text without switching grammar.
2. sub-word level LSTM (Prabhu et al., 2016): They passed character embeddings through 1-D convolutions and generated sub-word level representations. This was maxpooled and fed into the LSTM layer.
3. Hierarchical LSTM (Santosh & Aravind, 2019): They first generated phenomic sub-words, then trained a hierarchical LSTM model with attention on those words for classification.
4. CNN-1D (Kamble & Joshi, 2018): This work introduced domain-

specific word embeddings and reported their results on CNN, LSTM, and BiLSTM models. The CNN performs better than the rest.

### *6.2.3. HOT Baselines*

1. Word2Vec + FastText + [SS + LIWC + PV] (Mathur et al., 2018): They proposed using the Multi-Channel Transfer Learning model on word embeddings as input along with secondary input features that are Sentiment Score, Linguistic Inquiry+ Word Count, and Profanity Vector.
2. Debiasing and Author Profiling Deep Neural Network (Chopra et al., 2020): They used a bias elimination algorithm, graph embeddings with a concatenated profanity score, and passed this into convolutional and biLSTM networks with attention.

The proposed work shows an average increase of 7% for the TRAC-I dataset, 4% for the HS dataset and 1% for the HOT dataset in F1 scores. We investigated the results and predictions to formulate reasonings for the results in Section 7.

## *6.3. Comparing with Data simulations using Indic-Trans*

This experimentation aims to record the performance of the existing publicly available character-level transliteration library, Indic-Trans (Bhat et al., 2015) with the MuRIL model, and compare it to our word-based transliteration. We aim to process raw data using Indic-Trans library (Character-level transliteration), and compare it with *MoH* pipeline. We created three simulations of all datasets for extensive experimentation, shown in Table 10.

Table 9: Comparison with the state-of-the-art baselines for each dataset. '-' represents the unavailability of the metric score. The most significant results are in bold.

| Dataset | Technique | Precision | Recall | F1 |
|---|---|---|---|---|
| TRAC -I | M-Bert (Modha et al., 2020) | 0.65 | 0.62 | 0.62 |
| | n-gram + SVM (Kumar et al., 2018a) | - | - | 0.64 |
| | Dense NN (Raiyani et al., 2018) | - | - | 0.59 |
| | FeatureVec + LR(Samghabadi et al., 2018) | - | - | 0.64 |
| | fastText + CNN (Modha et al., 2018) | 0.63 | 0.62 | 0.62 |
| | Feature extraction + CNN (Singh et al., 2018) | 0.57 | 0.59 | 0.58 |
| | ***MoH* + M-Bert** | **0.74** | **0.60** | **0.66** |
| | ***MoH* + MuRIL** | **0.79** | **0.67** | **0.71** |
| HS | Synthetic + Gold (Samanta et al., 2019) | 0.59 | 0.63 | 0.53 |
| | Sub-Word Level LSTM (Joshi et al., 2016) | - | 0.36 | 0.46 |
| | Hierarchial LSTM (Santosh & Aravind, 2019) | - | 0.45 | 0.49 |
| | CNN-1D (Kamble & Joshi, 2018) | 0.83 | 0.78 | 0.80 |
| | ***MoH* + M-Bert** | **0.83** | **0.80** | **0.81** |
| | ***MoH* + MuRIL** | **0.86** | **0.84** | **0.85** |
| HOT | Word2Vec + FastText + [SS + LIWC + PV] (Mathur et al., 2018) | 0.816 | 0.92 | 0.89 |
| | Debiasing and Author Profiling DNN (Chopra et al., 2020) | - | - | 0.77 |
| | ***MoH* + M-Bert** | **0.84** | **0.93** | **0.88** |
| | ***MoH* + MuRIL** | **0.87** | **0.95** | **0.90** |

A step by step variation in each simulation is shown in Figure 5. Clear examples of these simulations are given in Table 11. These simulations are as follows:

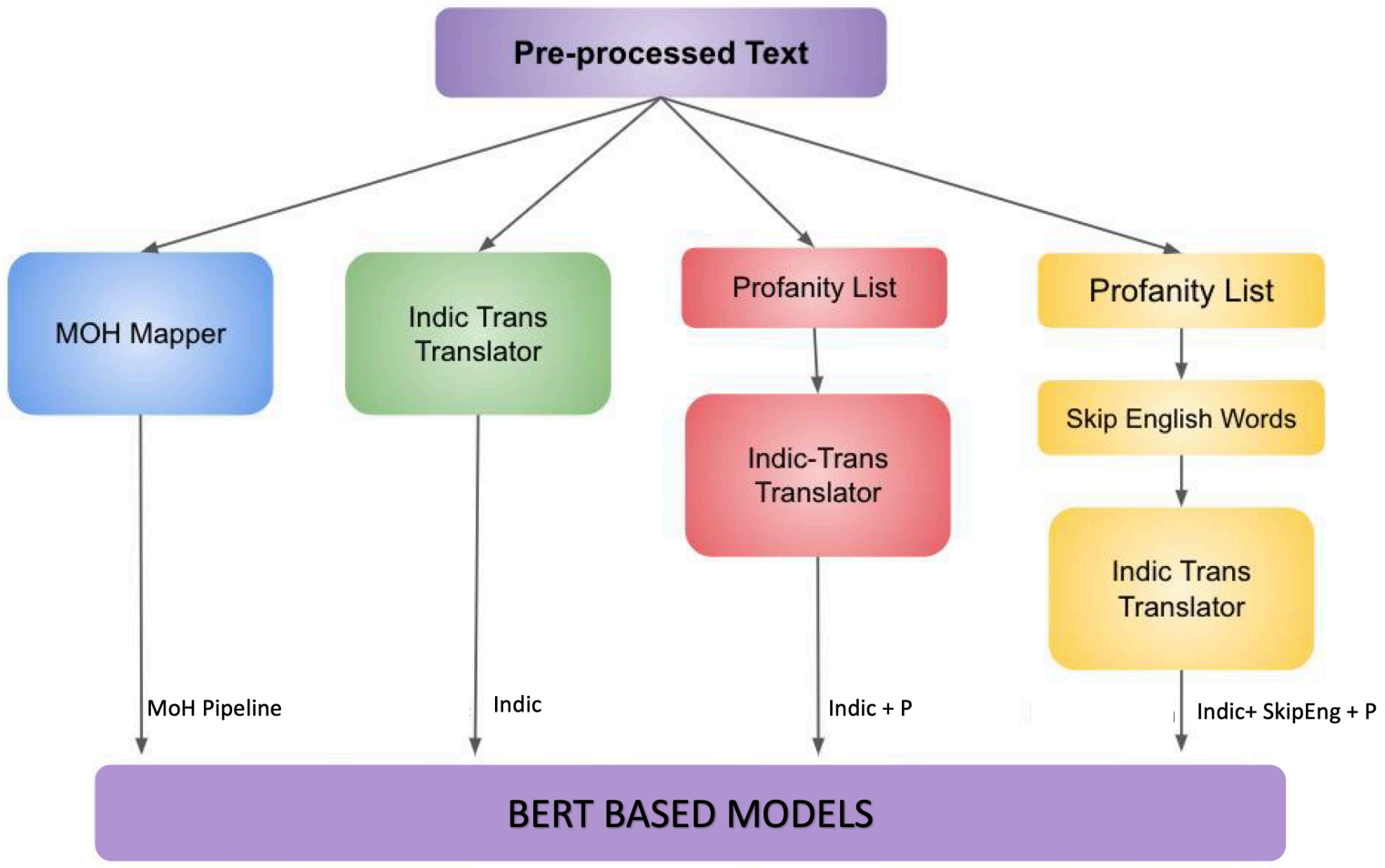

Figure 5: Types of data simulations

1. **Using only Indic-Trans (Indic)**: In this simulation, the Indic-Trans library translates the entire text to Devanagari Hindi. This means it converts even the English and Roman Hindi words into Devanagari Hindi.

Table 10: Comparison of simulations formed using indic-trans with proposed *MoH* + M-Bert

| Dataset | Model | Precision | | | Recall | | | F1 | | |
|---|---|---|---|---|---|---|---|---|---|---|
| | | Class 0 | Class 1 | Class 2 | Class 0 | Class 1 | Class 2 | Class 0 | Class 1 | Class 2 |
| TRAC-I | ***MoH* + MuRIL** | **0.78** | **0.82** | **0.75** | **0.74** | **0.64** | **0.63** | **0.72** | **0.73** | **0.66** |
| | Indic + MuRIL | 0.53 | 0.73 | 0.67 | 0.66 | 0.63 | 0.59 | 0.58 | 0.67 | 0.62 |
| | Indic + P + MuRIL | 0.69 | 0.75 | 0.67 | 0.67 | 0.56 | 0.60 | 0.67 | 0.64 | 0.63 |
| | Indic+Skip_En+P+MuRIL | 0.72 | 0.81 | 0.71 | 0.73 | 0.56 | 0.59 | 0.66 | 0.69 | 0.63 |
| HS | ***MoH* + MuRIL** | **0.85** | **0.87** | - | **0.82** | **0.85** | - | **0.85** | **0.85** | - |
| | Indic + MuRIL | 0.57 | 0.76 | - | 0.76 | 0.71 | - | 0.65 | 0.73 | - |
| | Indic + P + MuRIL | 0.57 | 0.77 | - | 0.76 | 0.74 | - | 0.65 | 0.75 | - |
| | Indic+Skip_En+P+MuRIL | 0.82 | 0.82 | - | 0.81 | 0.80 | - | 0.80 | 0.81 | - |
| HOT | ***MoH* + MuRIL** | **0.90** | **0.86** | **0.76** | **0.95** | **0.95** | **0.95** | **0.93** | **0.92** | **0.87** |
| | Indic + MuRIL | 0.72 | 0.76 | 0.66 | 0.80 | 0.82 | 0.74 | 0.75 | 0.78 | 0.78 |
| | Indic + P + MuRIL | 0.75 | 0.77 | 0.67 | 0.83 | 0.82 | 0.74 | 0.79 | 0.79 | 0.78 |
| | Indic+Skip_En+P+MuRIL | 0.84 | 0.85 | 0.76 | 0.81 | 0.89 | 0.83 | 0.86 | 0.83 | 0.82 |

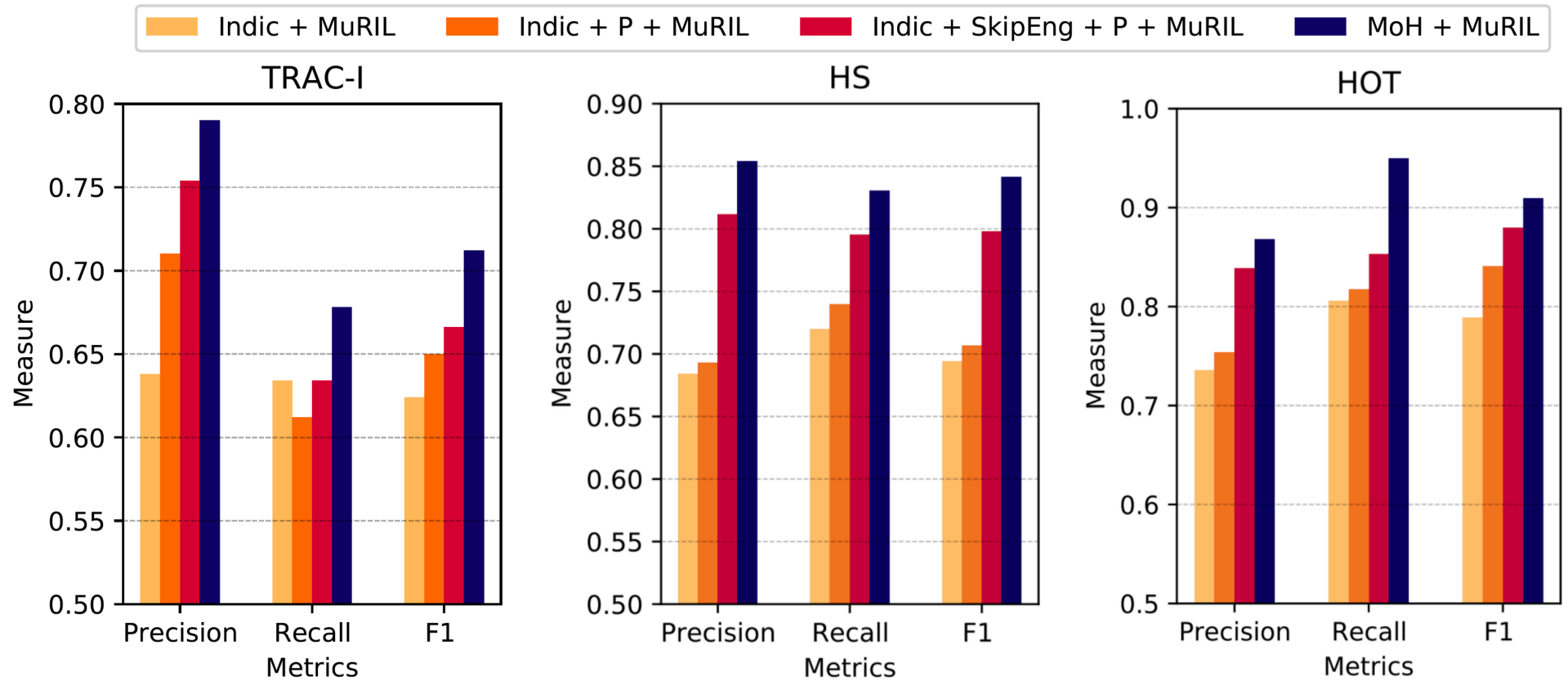

Figure 6: Comparison of Cumulative Evaluation Metrics for all the variations.

Table 11: Comparison between Data Simulations on an Offensive tweet. *The wrongly transliterated words are marked in blue.*

| | |
|---|---|
| Tweet | ramu suchha journalist h haramkor nahi |
| Reference | Ramu is an honest journalist not corrupt |
| *MoH* pipeline | रामू सच्चा journalist हैं हरामखोर नहीं |
| Indic | रामू सुचा जर्नलिस्ट ह हर्मकोर नहीं |
| Indic + P | रामू सुचा जर्नलिसट ह हरामखोर नहीं |
| Indic + Skip_Eng + P | रामू सुचा journalist ह हरामखोर नहीं |

2. **Translating Profane words separately (Indic + P)**: We observed that Indic-Trans's character-level translation incorrectly converted some Roman Hindi words into Devanagari Hindi forms. This is shown in Table 10. Since profane words play an essential role in detecting abusive speech, we decided to replace the profane words with their Devanagari Hindi translation using *MoH* Knowledge Base before running full text through Indic-trans. This ensured that the profane words were correctly translated into Devanagari Hindi.

3. **Translating Profane words separately and keeping English words intact (Indic + Skip_En + P)**: In the third simulation, we focused on eliminating Indic-Trans's character level translation on English words. The library produced meaningless Devanagari Hindi words by changing English letters into Devanagari Hindi. We performed language identification to keep the English words intact and then converted profane Roman Hindi words into Devanagari Hindi using *MoH* Knowledge Base. The resulting data is a combination of Devanagari Hindi and English text.

From Table 10 and Figure 6, we see that for all the datasets, our proposed pipeline generally outperforms these simulations. MoH + MuRIL shows an increase over Indic + MuRIL of about 33%. This happens because character-level translation can lead to severe spelling errors rendering the translated word meaningless. In the ‘Indic + MuRIL’ technique, Indic-Trans converted English words and Roman Hindi profane words token-wise, giving incorrect spellings. The tokenizer may not have captured these words.

In the second and third simulations, ‘Indic + P’ and ‘Indic + Skip_En + P’, we eliminated the effect of incorrect conversion of English and Profane words. Yet, word-based *MoH* mapping technique outperformed Indic-Trans. We infer this is due to the character-level transliteration of Roman Hindi words to Devanagari Hindi, which is erroneous because the character level transliterator does not check if its output words exist within the Hindi language. Our pipeline easily avoids this obstacle.

The clear distinction between transliterations using Indic-Trans and *MoH* pipeline are shown in Table 11 with example word *‘corrupt’* (written as हरा-मखोर in Devanagari Hindi and **haramkor** in Roman Hindi). Indic-Trans blindly transliterates individual letters of the English word into Devanagari Hindi, i.e., हर्मकोर. Moreover, English words like *‘journalist’* are incorrectly transliterated to जर्नलिस्ट. These words are meaningless in Devanagari Hindi.

### *6.4. Error Analysis*

To understand the performance errors, we conducted a qualitative error analysis, focusing on the false negatives (FN), and false positives (FP), of our work, for each of the three datasets.

Our model misclassified 311 posts of the TRAC-I data, 192 posts of the HS data and 88 posts of the HOT data. It is evident from Table 12 that the number of false positives and negatives for each dataset is adequately low for both Non-Offensive and Offensive posts (OAG and CAG classes of TRAC-I, and Abusive and Offensive classes of HOT). For Trac-I, and HOT datasets, the percentage of false positives is slightly higher than the percentage of false negatives, while the percentages are comparable for the Offensive class in the HS dataset.

Table 12: Table depicts the distribution of test set, misclassified posts in the test set and the percentage of False Positives (FP), False Negatives (FN) out of the total Non-Offensive and Offensive posts for each of the datasets.

| | TRAC-I | HS | HOT | Total |
|---|---|---|---|---|
| No. of Total Posts | 971 | 915 | 736 | 2622 |
| No. of Misclassified | 311 | 192 | 88 | 591 |
| % FP - Non Offensive | 25.54% | 9.81% | 12.3% | 13.2% |
| % FN - Non Offensive | 13.02% | 7.08% | 9.23% | 8.46% |
| % FP - Offensive | 17.4% | 15.09% | 6.55% | 12.82% |
| % FN - Offensive | 13.08% | 15.84% | 4.47% | 10.15% |

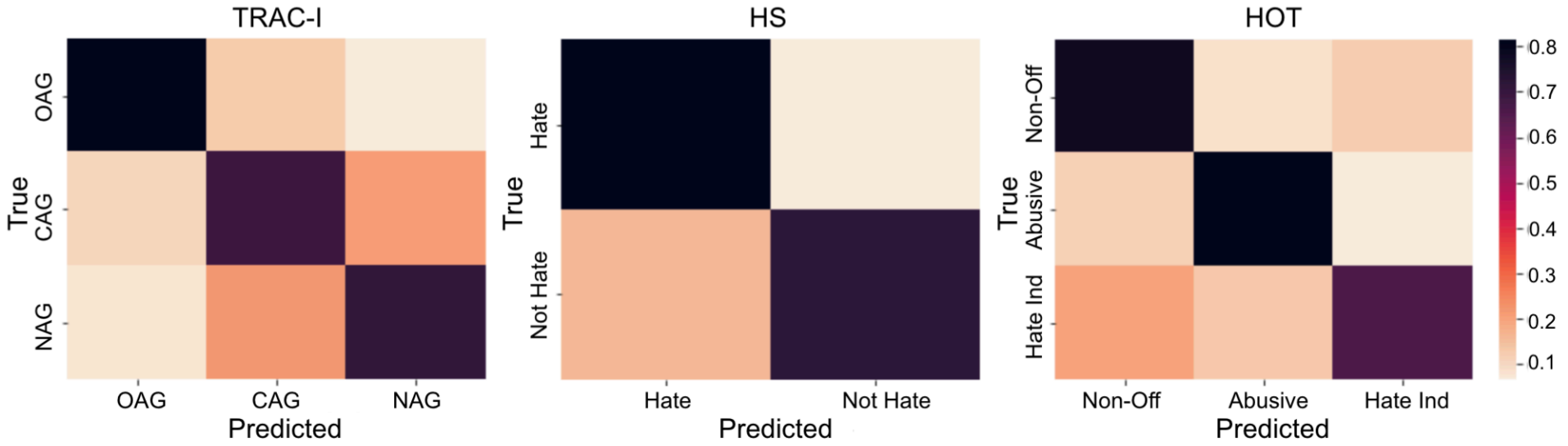

Figure 7: Normalized Confusion Matrix of TRAC-I, HS and HOT datasets respectively. In Trac-I, OAG, CAG and NAG represent Overtly, Covertly and Non Aggressive classes. In HOT, Non-Off, Abusive and Hate Ind represent Non-Offensive, Abusive and Hate Inducing classes.

Figure 7 shows that the maximum false positives in the first plot arise from 'Covertly Aggressive' texts that are falsely classified as 'Non Aggressive' and vice versa. The 'Overtly Aggressive' posts have been predicted better than the other two classes. The second plot shows that the 'Hate Speech' posts have a higher percentage of false positives compared to the 'Normal Speech' posts. In the third plot, the 'Non-offensive' and 'Hate Inducing' have relatively more significant false positives than 'Abusive'.

### *6.5. Time Analysis*

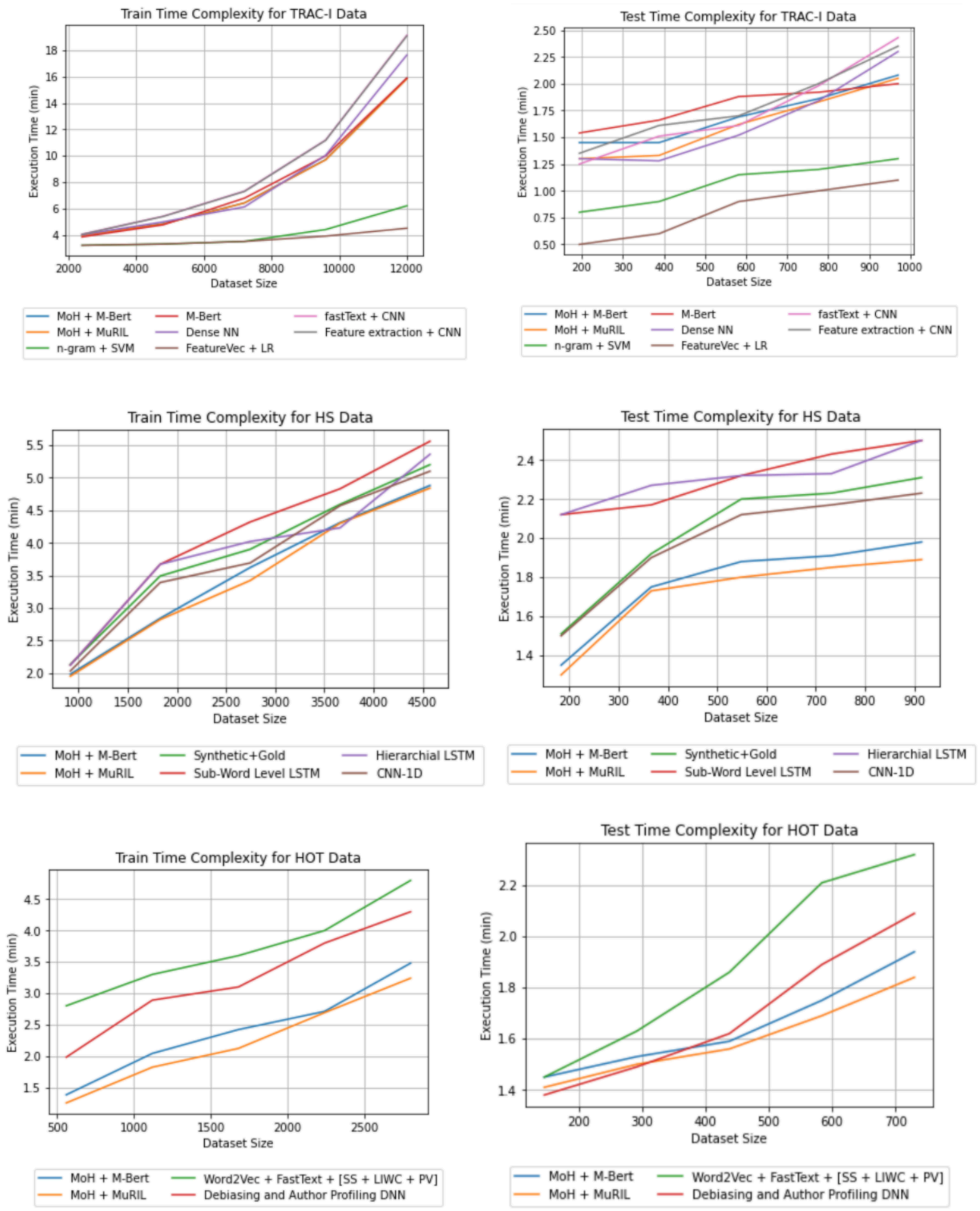

Figure 8: Train and Test run time complexity for TRAC-I, HS and HOT data respectively.

This section compares the run time complexity performance of the proposed work on the three datasets. Figure 8 shows the training and testing time for TRAC-I, HS, and HOT data, respectively. We compare the proposed method with the baselines shown in Table 9 for each dataset. For both training and testing data, we infer that the computational complexity of LSTM and CNN models have contributed to an increase in their running time (Vaswani et al., 2017). However, we observe that transfer learning methods such as fine-tuned pre-trained Bert based models have reduced latency. Machine Learning models such as LR and SVM show visibly lower complexities than deep learning strategies. We conclude that MOH + M-Bert and MOH + MuRIL performs at par with most of the baselines.

## 7. Discussion

We used the statistics in Section 6.4 to formulate inferences and address the errors in our model predictions. We collected the text samples of misclassified data, shed light on finding the reasons for the incorrect classifications with adequate analysis, and support our inferences with substantive examples from the datasets. We observed the common traits between misclassified texts to form intuitive correlations and relied on empirical scoring to confirm it. Making the model robust to these errors requires further investigations and falls out of the scope of our work.

*7.1. Unawareness of Context*

We came across sentences that used cultural or social references to attack specific individuals (communities). Unless there is not enough contextual data, the model cannot disambiguate the meaning of these posts. We noticed this in politically inclined posts that are present in TRAC-I data. A few example posts from TRAC-I data are *'modi ne le lia sehjade ka hahhaha'* (English translation: Modi defeated the Prince). Here, *sehjade* refers to Rahul Gandhi, Indian political leader, and the term means 'prince'. Although the term itself is not derogatory, it has often been used in media to attack Rahul Gandhi due to the dynastic politics that governs his party.

Another post is *sahi kaha modi mogambo h*, which is labeled 'Covertly Aggressive' since *'mogambo'* is a popular Hindi movie's fictional villain. The post wants to convey that 'Modi is like a villian' but means 'correctly said Modi is Mogambo'. In our view, these posts needed context words or worldly knowledge to classify it as hate speech.

*7.2. Implicit hate*

TRAC-I data uses certain parts of speech that may be regarded as hateful even if the individual words by themselves are not offensive. Example: *'modi gujraati school se haha'* (English translation: Modi is from gujarati school *laughs*) does not include hateful words; however, the sentence is mocking the subject's educational background, and so this post was labelled as Covertly Aggressive. Another post is *'ye hain achche din?'*(English translation: Are these the good days?), which criticizes the Indian political party's promise of 'achche din aane waale hain'(English translation: Good days are coming soon). The models may have been confused due to ‘achche’, which means ‘good’, and misclassified this post. Implicit hate also occurs when two-sided pronouns that combine the in-group and out-group (e.g., your/our, you/us, they/we) and the pronoun patterns such as verb-pronoun combinations, which capture the context in which two-sided pronouns, are used (e.g., send/them, protect/us) (Burnap et al., 2014; Alorainy et al., 2019). For instance: *‘send them to jail’.* Our classifier did not capture these linguistic features.

### *7.3. Profanity classified as hate*

Sentences that comprise profane words are not necessarily hate-inducing, as shown in Table 2. However, our model may not have understood the connotations of such profanity when the text does contain profane words but does not inculcate hatred. Example: *mohit vaise toh haramkhor, lekin padhne mein hoshiyar hai* (English translation: Mohit is generally shrewd, but he studies hard). In this post, the profane word is ‘haramkhor’, however, the sentence highlights Mohit’s positive quality. Our model labelled this post as ‘hate’. Multiple instances of such type of post are present throughout all three datasets.

### *7.4. Dubious Annotation*

TRAC-I, HS and HOT datasets annotations were carried out by 1, 2, and 3 individuals, respectively. The inter-annotator agreement was reported as a 0.83 kappa score for the HOT dataset and a 0.982 kappa score for the HS dataset. However, on manual inspection of some of the false positive and false negative posts, we observed that even the true labels were incongruent with our perception of the posts. This was observed in all three datasets. In HOT data, tweets such as *'ki gadha nahi'*(English translation: not a donkey) and *'excuse my potty mouth'* are labelled as 'hate Inducing', when they do not imply any sense of hate. Similarly, in HS data, *'hate it jab test ata ho phr bh acha na ho'* (English translation: hate it when even after preparation, test does not go well) and *'college se sakht nafrat hai. Mujhe 10th ki chuttiyan wapis la do'* (English translation: I don't like college, give my 10th class holidays back) are incorrectly labelled as 'hate speech'. These posts are not attacking or discriminating against individuals or communities, yet they were labelled as hate speech.

### *7.5. Short Posts / Sarcasm / Rhetorical Questions / Metaphors*

After pre-processing, specific posts in the datasets were concise sentences which either had misspelt words or formed grammatically incomplete text (2 words posts). Example: *'dogi nigam'*, *'potty turns'*, *'still hate'* in HOT Dataset are labelled as hate inducing. Sarcasm or irony appeared in most of the TRAC-I data's Covertly Aggressive labelled samples. Since sarcasm uses words to imply an opposite meaning, our classifier cannot accurately spot such posts without prior knowledge. Example post: *'Aur jitao modi ko'* and *'bhasha toh seekh lo'* is mocking and sarcastically exclaiming 'Keep voting for Modi' and 'At least learn the language' respectively. Some of the posts were virulent and suggestive rhetorical questions such as 'Yehi hai acche din?' (English translation: Are these the good days?), questioning the Indian political party's political campaign. Online social platforms have plenty of such comments. To understand such styles of speech, classifiers need awareness about the world or language in depth.

## 8. Conclusion and Future Works

In this work, we investigated the problem of code-switching text in the domain of hate speech detection. We proposed a word-based transliteration pipeline, *'MoH'*, to convert all Roman Hindi words correctly into Devanagari Hindi script despite the vocabulary challenges of Roman Hindi while keeping the English words intact. Moreover, we fine-tuned Bert-based models, M-Bert and MuRIL, to leverage their pretrained multilingual capabilities for classification. Our system offers excellent potential for improved performance. It is backed by extensive experimentation, comparison with existing techniques, and in-depth error analysis of our model's predictions. In future works, we plan to enhance the model to tackle all the error-causing factors mentioned in Section 7, as these are ubiquitous in social media platforms.